# Segmentation Approach for Coreference Resolution Task

Aref Jafari        Ali Ghodsi

University of Waterloo

{aref.jafari, ali.ghodsi}@uwaterloo.ca

**Abstract**

In coreference resolution it is important to consider all members of a coreference cluster and decide about all of them at once. This technique can help to avoid losing precision and also in finding long-distance relations. The presented paper is a report of an ongoing study on an idea which proposes a new approach for coreference resolution which can resolve all coreference mentions to a given mention in the document in one pass. This has been accomplished by defining an embedding method for the position of all members of a coreference cluster in a document and resolving all of them for a given mention. In the proposed method, the BERT model has been used for encoding the documents and a head network designed to capture the relations between the embedded tokens. These are then converted to the proposed span position embedding matrix which embeds the position of all coreference mentions in the document. We tested this idea on CoNLL 2012 dataset and although the preliminary results from this method do not quite meet the state-of-the-art results, they are promising and they can capture features like long-distance relations better than the other approaches.

## 1   Introduction

Over the past decades several models have been proposed for the coreference resolution task[1]. Among all of them, the machine learning-based models can be categorized into three major categories: mention pair models, mention ranking models, and entity-mention models[2]. In all these categories, the mentions first detect and then the models find the coreference relation between them or between mentions and an entity. In contrast to this idea, we believe the problem of coreference resolution can be viewed as a segmentation problem. Using the concept of image segmentation problems where we classify the related pixels of a specific object in an image, here we can segment the position of the tokens in a given document as the coreference mentions. Each mention in a document consists of one or more connected tokens; therefore, they have start and end token indices. One way that coreference resolution can be defined is predicting all start and end indices of mentions in a coreference cluster. To design a model for this task, we first need a good embedding method to encode the document in a context space with consideration for the relations between words and tokens, and then we can design a decoder to decode the embedded context vectors of the document into the indices of coreference mentions. Over the past few years, we have witnessed the revolution of transformer-based[3] models in the natural language processing world. The new generation of pre-trained encoder models like BERT[4], XLnet[5] and other similar models have the ability to embed the input document by considering the relation between all tokens in it. Therefore, they can encode the tokens of a sentence into a common context space by considering all important relations between the tokens. By using these encoding models, we can design a decoder to decode the embedding context of a document and find the position of related mentions in a cluster. Based on this idea, we can define the coreference resolution task as the task of finding all coreference mentions to the given mention in the given document.

To the best knowledge of the authors, most of the previous coreference resolution models are similar in explicitly extracting the name entities and then finding the coreference relations. In this paradigm, we lose much of the useful information in the context as well as the integrity between tokens. Therefore, these models can not have good exploitation over the context information in the document.

Although recent end-to-end models like[6][7][8] try to address some of these problems, they cannot solve them completely since they are still using mention ranking.

## 2 Related Works

Coreference resolution is a classic problem which has been approached over the past few decades by numerous researchers[1]. These researchers have tried to solve this problem by using the most advanced methods they have in their era including some of the famous categories such as linguistic, rule-based, graphical model, and neural network-based methods. Presently, the neural network-based models have achieved the best results as compared to other approaches. After several attempts of applying deep neural networks to this problem, the most successful end-to-end neural network-based method was proposed by Kenton Lee et. al. in 2017 for the first time[9]. Later, other similar methods had been proposed and these improved the previous results[7][8][10]. It would be reasonable to say that most of the modern coreference resolution models expand all possible spans up to a certain length and learn a conditional distribution over them where the most likely configuration of this distribution produces the correct clustering of the mentions. Basically, these models have two scorer functions - one for scoring whether a candidate span is a good mention or not, and the other for scoring if two given mentions are coreferenced to each other or not.

These two processes have been integrated into a unified end-to-end model and for exploiting context information a sequence embedding network, like bidirectional LSTM, is used to embed the tokens. The consideration of all possible spans is a time-consuming process which can be eliminated with a simple scheme. In our proposed approach we tried to remove the explicit span detection phase and let the model find the coreference mentions at the output. We believe that this approach model can leverage the context of the document to detect the correct coreference relations between the mentions.

## 3 Proposed method

Unlike other approaches which decide the individual mentions, we propose a model to detect all coreference mentions of a cluster implicitly in one pass. For this purpose, we gain inspiration from the "question answering" approach used in the BERT model[4]. In the question answering model of BERT, given a question and a paragraph to the model, we train two query vectors on top of the BERT model to detect the start and end tokens of the answer in the given paragraph. We can use the same idea, and by giving the document and one of the mentions to the BERT

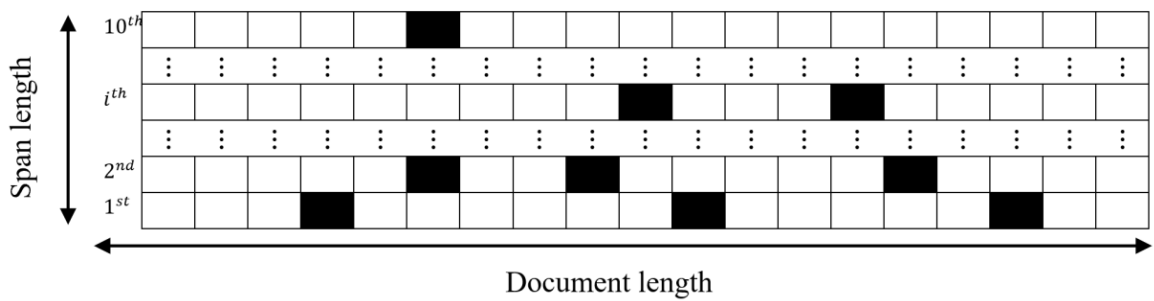

**Figure 1.** Span position embedding matrix. Each row in this matrix indicated the length of spans and each column indicates the position of the first token of the span

model we train a head network to detect the position of all mentions in the given document. For this purpose, we first need to define an encoding method for the position of the coreference mentions in a document. It is important to consider the fact that the mentions are spans with connected tokens. Also, we know from previous studies[9] that the coreference mentions with a length of more than ten tokens are rare. Therefore, we can define an encoding for the position of all possible spans in the given document up to ten tokens. For this purpose, we can consider a matrix $D$ with $k$ rows and $l$ columns where $k$ is the maximum length for possible spans and $l$ is the maximum token length in the input document. Each element $d_{ij}$ in this matrix shows the position of a span with length $j$, where an index $i$ of the given document begins. For example, the element in row 2 and column 4 indicates one of the spans with token indices 4 and 5 in the document or the element in row 4 and column 6 indicates a span with token indices 4, 5, 6, and 7. For the spans which belong to the related mentions with the input mention, we can set their position values to 1 and leave the rest span positions to be zero. This matrix can embed the position of all coreference mentions with the given input mention. Figure 1 illustrates this idea visually.

After defining the embedding method, the next step will be predicting this embedding matrix for a given document and mention. This is very similar to the segmentation task in computer vision. Instead of segmenting the related pixels of an object, we want to segment the position of mentions in each document which all refer to a given mention. For this purpose, we feed the BERT model with one of the mentions and the document in the same way for question answering task in this model[4].

[CLS] mention [SEP] Document [SEP]

Then, after getting the embedded outputs of tokens, we design a simple convolutional head network on top of the BERT to map these embedding vectors to the position embedding matrix of coreference mentions. As it is illustrated in figure 2, multiple convolutional kernels with different sizes have been used for mapping the output embedding of BERT to the defined span position embedding matrix. If we consider that each embedding vector of tokens in BERT has $m$ dimension, then each convolutional kernel with size $m \times i$ can be used to reconstruct the $i^{th}$ row of the span position embedding matrix. In the end, we stack the resulted vectors to reconstruct the span position embedding matrix. The reason for using convolutional kernel with width $i$ for reconstructing the $i^{th}$ row of the span position embedding matrix is that the $i^{th}$ row of this matrix embeds the positions of the spans with length $i$, therefore a kernel with size $m \times i$ can capture the necessary information about the relations between the tokens of these spans to determine whether this span can be an appropriate coreference mention for the given input mention or not. Note that the relations between all tokens of the document have already been captured by attention layers of BERT before, and this information is encoded in the output embedding vectors of BERT. So in this way, the BERT model encodes the relations between all tokens in the document in each embedding vector of tokens, and the convolutional head captures the needed information from the spans' tokens to decide whether they are a coreference or not.

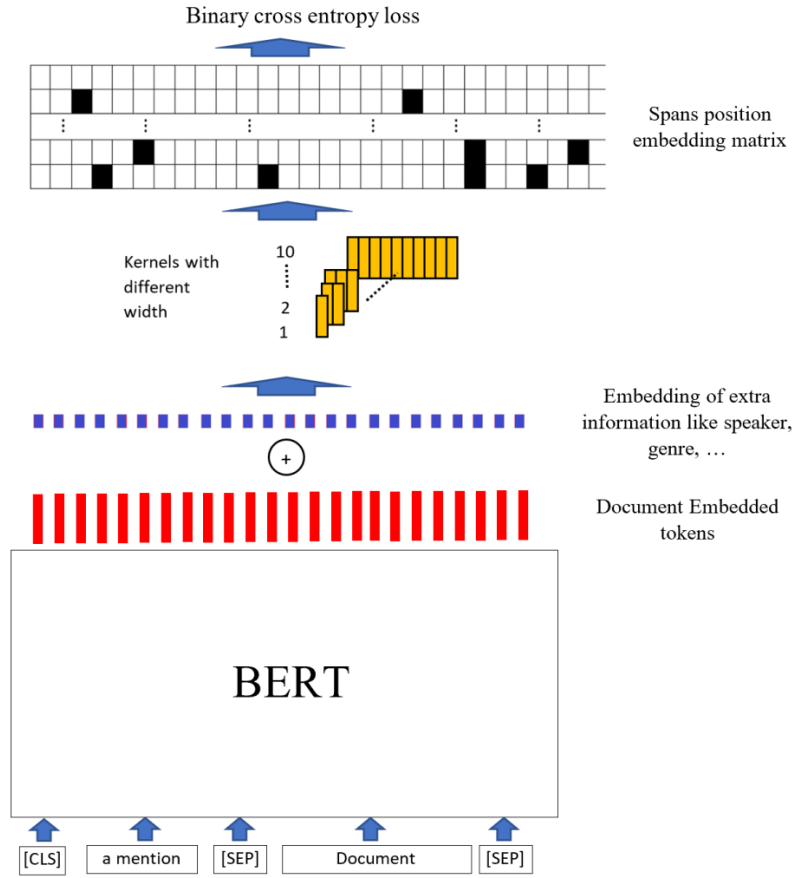

**Figure 2.** Structure of the proposed method

We used the binary cross-entropy loss function to fit the output of the network to the span position embedding of the given mention during the training time. Also, for CoNLL 2012 data we embed metadata such as speaker and genre information as one bit and three bits binary feature and concatenate them to the BERT output embedding vectors. The convolutional kernels have been applied on the resulted vectors.

|  | $B^3$ | | | MUC | | | CEAF | | | Avg. F1 |
|---|---|---|---|---|---|---|---|---|---|---|
|  | precision | recall | F1 | precision | recall | F1 | precision | recall | F1 |  |
| Proposed method | 80.4 | 55.9 | 65.9 | 84.6 | 60.3 | 70.4 | 80.5 | 39.4 | 52.8 | 63.0 |
| State-of-the-art | 76.5 | 74.0 | 75.3 | 84.7 | 82.4 | 83.5 | 74.1 | 69.8 | 71.9 | 76.9 |

**Table 1.** Results

## 4 Experiments

We used CoNLL 2012 dataset[11] to test this approach. It is one of the most challenging datasets for the coreference resolution task. It consists of English, Chinese, and Arabic languages where we just focused on English data. The dataset is divided into train, test, and development parts. We used PyTorch implementation of BERT based model[12] and set the max length of input to 256 tokens (10 tokens for the given mention, 3 tokens for [CLS] and [SEP] and 243 tokens for the given document). We also encode the genre and speaker metadata. During this process, the head network trained and the BERT network fine-tuned at the same time. The batch size was set to 16 and the model was trained on an Nvidia tesla P40 graphic card. During the training each time one of the coreference mentions of the train set was fed to the network and the whole cluster of coreference mentions of the given mention was embedded as span position embedding matrix. We used both singleton and coreference mentions during training randomly. Also, for the documents longer than 243 tokens, we shrinked them into several segments with length 243 and with 50% overlap. For test time we trained a second network with the same structure of the main model to detect all mentions first. We only fed the document tokens to this network and set the position of all mentions in the span position embedding matrix to 1. We trained this network on the train set first and used it during the test time. This auxiliary network could detect all mentions of a document. We then chose the first detected mention and fed it into the main trained model to detect the coreference mentions with the given mention. After finding the first cluster, all members of that first cluster were removed from the found mentions set and then the next first mention from the set has been chosen to fed to the main model. We do this process for each document until all coreference clusters of the document have been detected.

The preliminary results of this method do not match the state-of-the-art results yet. They are nevertheless very promising and show the potential effectiveness of this approach. Table 1 illustrates our preliminary results with this model. As it illustrates, the method can achieve a good precision rate in all three scores although the model recall needs improvement.